\documentclass[letterpaper, 10 pt, conference]{ieeeconf}  
\usepackage[utf8]{inputenc}
\usepackage{siunitx}
\usepackage{graphicx}
\usepackage{amsmath}
\usepackage{amsfonts,amssymb}
\usepackage{hyperref}
\usepackage{booktabs}
\usepackage{subcaption}

\DeclareSIUnit\px{px}
\newcommand{\figurename}{Fig.}
 
\IEEEoverridecommandlockouts                              

\title{\LARGE \bf Metric Pose Estimation for Human-Machine Interaction Using Monocular Vision}

\author{Christoph Heindl$^{*}$, Markus Ikeda$^{*}$, Gernot St\"ubl$^{*}$, Andreas Pichler$^{*}$ and Josef Scharinger$^{**}$
\thanks{$^{*}$Visual Computing and Robotics, PROFACTOR GmbH, Austria
	{\tt\small christoph.heindl@profactor.at}}%
\thanks{$^{**}$Institute of Computational Perception, Johannes Kepler University, Austria
	{\tt\small josef.scharinger@jku.at}}%
}

\begin{document}

\maketitle
\thispagestyle{empty} 




\begin{abstract}
  The rapid growth of collaborative robotics in production requires new automation technologies that take human and machine equally into account. In this work, we describe a monocular camera based system to detect human-machine interactions from a bird's-eye perspective. Our system predicts poses of humans and robots from a single wide-angle color image. Even though our approach works on 2D color input, we lift the majority of detections to a metric 3D space. Our system merges pose information with predefined virtual sensors to coordinate human-machine interactions. We demonstrate the advantages of our system in three use cases.
\end{abstract}

\section{INTRODUCTION}
A central aspect of collaborative robotics is the reliable detection of human-machine interactions over large work spaces. Typical conventional sensors, such as light barriers and physical keys, do not provide the required level of perception or lead to non-intuitive operation. In addition, the integration of such sensors requires considerable amounts of time for rewiring and reprogramming \cite{brown2017operations}. 

Over the past years, several alternative sensor technologies have been proposed to close this gap: Tactile sensors provide reliable touch gesture detection \cite{silvera2015artificial}, whereas proximity sensors ensure human-robot safety aspects \cite{geiger2019160} in close vicinity of the robot. Recently, RGB-D sensors have been increasingly used to detect human poses \cite{fang2018understanding}. These sensors provide dense depth data, limited to a few meters distance. The light emitted by these cameras is easily scattered by shiny surfaces, rendering many measurements unusable.

\begin{figure}[t]  
  \centering
  \includegraphics[width=\columnwidth]{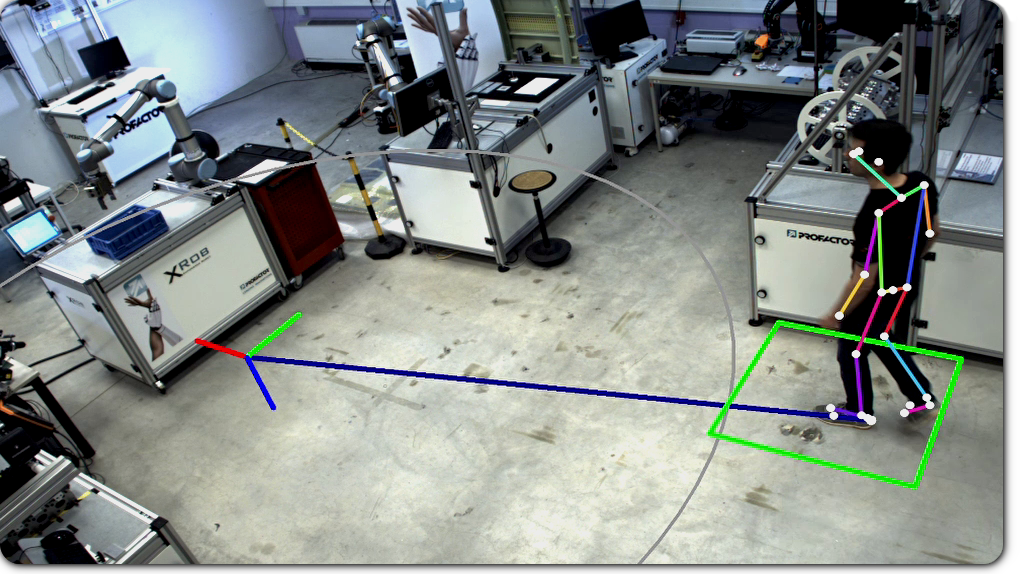}
  \caption{\label{fig:aeyeoverview} Our system coordinates human-machine interactions from a bird's-eye perspective. We introduce a set of extensible virtual regions that act as replacements for physical sensors. 2D human/robot poses are recognized from color input images and are lifted into a metric 3D space. The geometric relation between humans, robots and virtual regions enables our system to recognize interaction patterns and trigger environmental reactions.}
\end{figure}

Our system is illustrated in \figurename~\ref{fig:aeyeoverview}. We propose to replace a range of conventional hardware sensors by a single monocular vision system that operates from a bird's eye perspective. In contrast to RGB-D systems, our approach processes pure 2D color data from a wide-angle camera. Despite the consequent loss of depth, we show that 3D metric pose information is recoverable for humans and robots using homographies. We fuse pose estimation with an extensible set of virtual sensors to determine interaction events. We demonstrate the system in three use-cases involving multiple humans and robots\footnote{\label{fn:video}Demonstration video \href{https://youtu.be/BEqF-wFpxs0}{\url{https://youtu.be/BEqF-wFpxs0}}}.

\section{Method}
\begin{figure}[t]
  \centering
  \includegraphics[width=0.9\columnwidth]{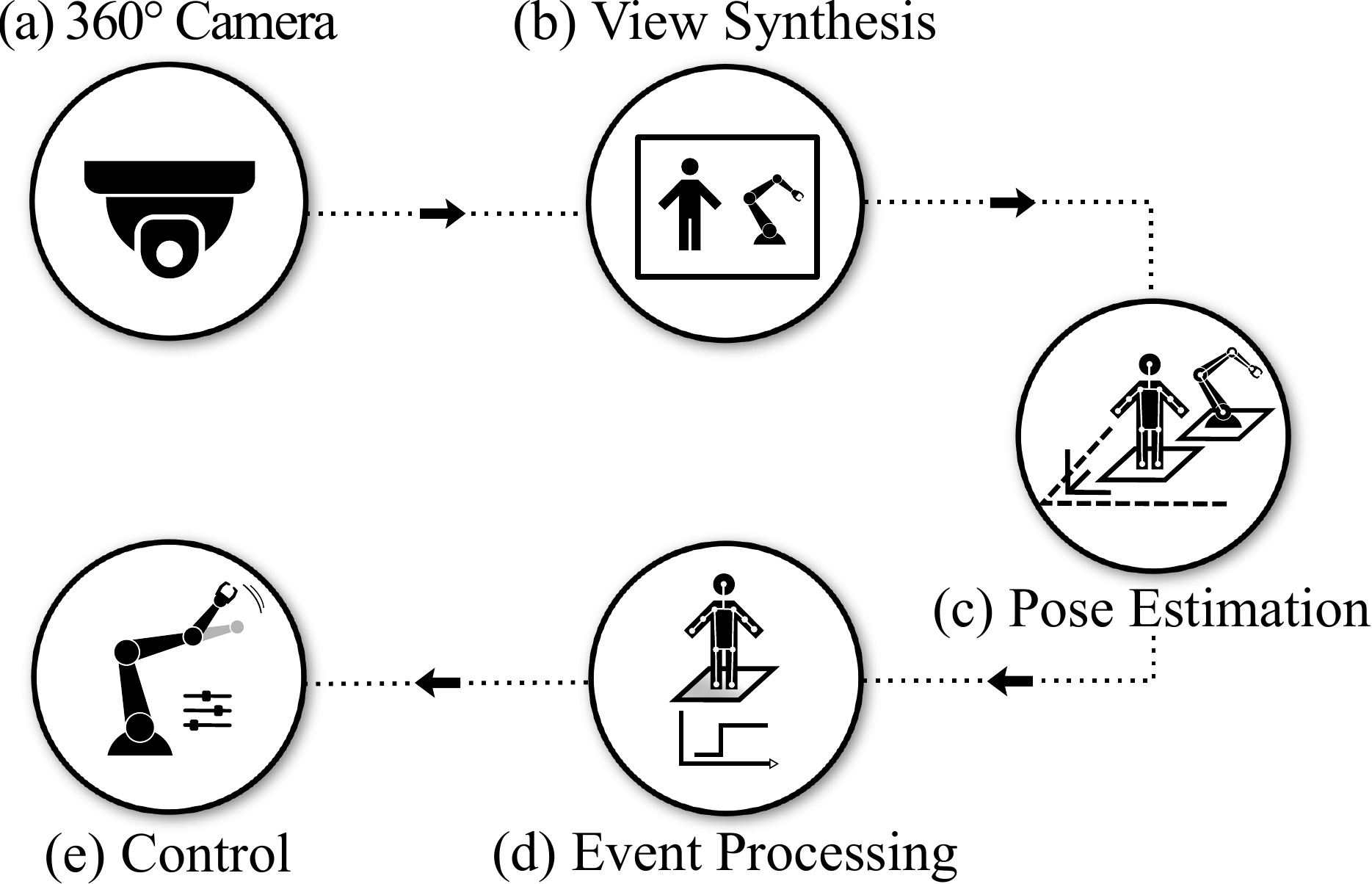}
  \caption{\label{fig:aeyegraph} System Overview. Our approach considers panoramic images as input (a); these are synthesized to form one or more rectilinear views (b). Deeply learned neural networks predict human and robot poses (c). A number of virtual regions raise location-aware events based on geometric relations to surrounding entities (d). These events in turn lead to application dependent environmental reactions (e).}
\end{figure}

The architecture of our approach is depicted in \figurename~\ref{fig:aeyegraph}. Our system first synthesizes rectilinear views \cite{heindl2018} from a single panoramic color image. In rectilinear views straight lines are preserved, so that our system is enabled to build on machine learning models particularly engineered for such optics. 

\begin{figure}
  \centering
  \includegraphics[width=0.95\columnwidth]{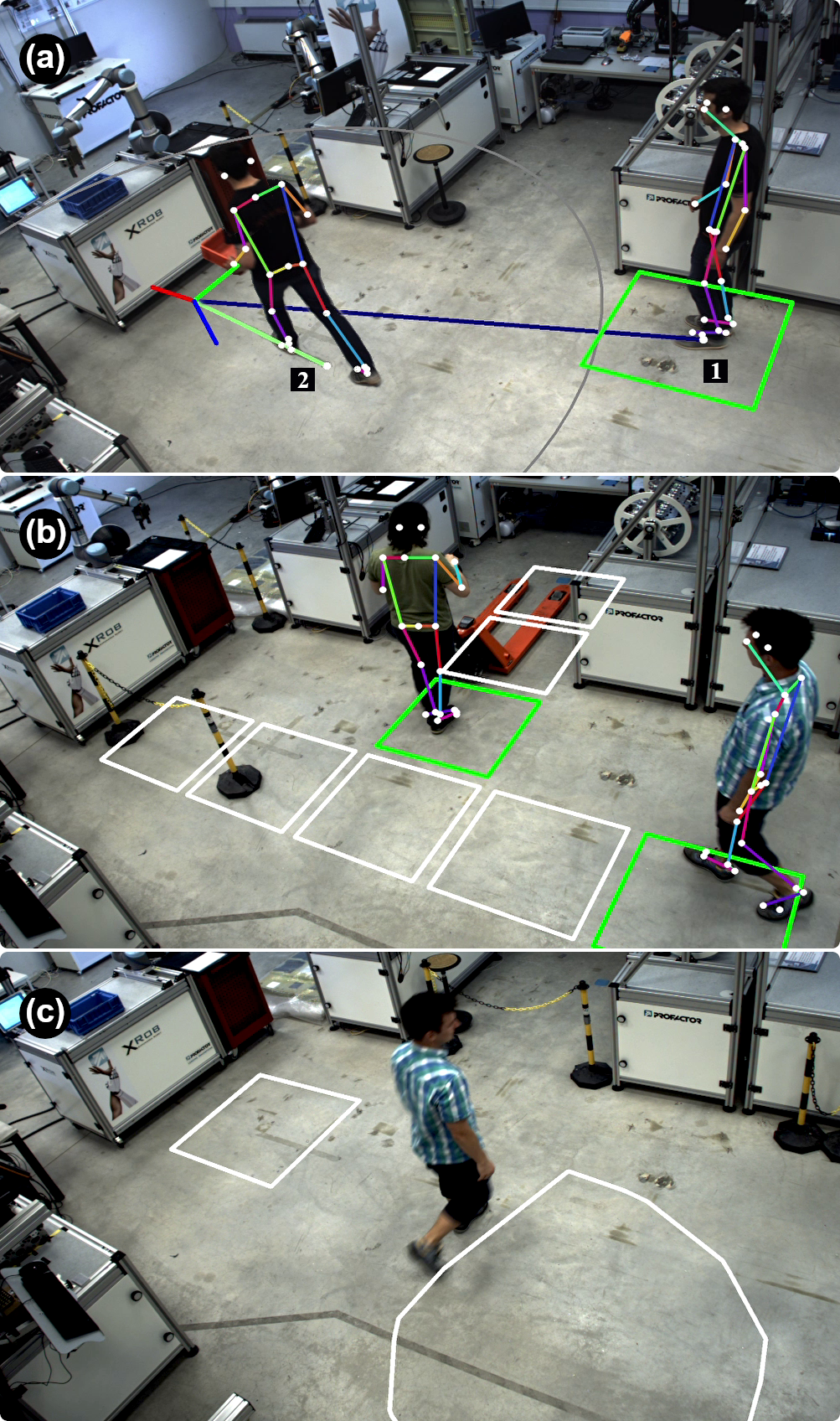}
  \caption{\label{fig:usecase} Use cases (UCs); (a.1) UC1 Entering the (green) rectangular sensor enables the robot to start; (a.2) the worker's proximity to the robot controls its speed. (b) UC2 Regions are only sensitive to humans, but are not influenced by other objects. (c) UC3 Free-form region definition by humans.}
\end{figure}

Next, we perform 2D human and robot pose estimation \cite{cao2017realtime, heindl2019learning} on each view (see \figurename~\ref{fig:usecase}a-c). These detections are then transformed into metric space by mapping pixel locations to world coordinates using plane homographies. In particular, a homography between ground and camera image plane allows us to convert pixel to metric ground coordinates. Since such a mapping is valid only for keypoints semantically close to the ground (such as feet positions), we additionally integrate statistical height measurements of an upright body model in order to map hip and shoulder keypoints. These extra points are used to predict body orientation and to stabilize localization in the presence of occlusions. 

Once world coordinates are established, our system scans for events that arise from the interaction of humans with a predefined set of virtual sensors placed in ground coordinates. Each such event might then trigger one or more environmental reactions, depending on the application. Among others, our system supports the following virtual sensors: light barriers, (freeform) step-on sensor mats, proximity sensors and body orientation aware sensors.
 
\section{Demonstrations}
We demonstrate the interaction potential of the AEYE system in three Use Cases (UCs)\footnote{See footnote~\ref{fn:video} for video link. }:
  
  \emph{UC1} focuses on human-robot interaction (see \figurename~\ref{fig:usecase}a). A human entering the rectangular region enables the robot to start. The robot's speed is adjusted according to the distance to the closest worker. This contributes to creating safe operating environments for the interaction between man and machine.

  In \emph{UC2} we demonstrate several people interacting with multiple regions. The focus is on system robustness in the presence of occlusion and the system's ability to react to people only---avoiding false triggers caused by other objects (see \figurename~\ref{fig:usecase}b).
  
  In \emph{UC3} we define virtual sensors by visual demonstration through human movements (see \figurename~\ref{fig:usecase}c). Teaching by demonstration significantly reduces the time required to create and program virtual regions. Note that action detection is currently not part of our system. The start and stop gestures are triggered by a wireless presenter in this use case.

\section{Conclusion}
We demonstrate that the challenging task of monitoring human-machine interactions on a metric scale is largely feasible using a single color camera. Combining the latest results of deep learning with traditional computer vision methods lifts many 2D recognitions to metric scales. This allows us to replace many physical sensors with virtual surrogates. 





\bibliographystyle{IEEEtran}
\bibliography{aeye_fof}								

\end{document}